\documentclass{article}

\usepackage[preprint]{neurips_2022}

\usepackage[utf8]{inputenc} 
\usepackage[T1]{fontenc}    
\usepackage{hyperref}       
\usepackage{url}            
\usepackage{booktabs}       
\usepackage{amsfonts}       
\usepackage{nicefrac}       
\usepackage{microtype}      
\usepackage{xcolor}         
\usepackage{amsthm}
\usepackage{svg}
\usepackage{caption}
\usepackage{subcaption}

\newtheorem{definition}{Definition}[section]
\newtheorem{theorem}{Theorem}[section]

\title{Topological Data Analysis of Neural Network Layer Representations}

\author{
    Archie Shahidullah \\
    Computing + Mathematical Sciences \\
    California Institute of Technology \\
    Pasadena, CA 91125 \\
    \texttt{archie@caltech.edu}
}

\begin{document}

\maketitle

\begin{abstract}
This paper is a cursory study on how topological features are preserved within the internal representations of neural network layers. Using techniques from topological data analysis, namely persistent homology, the topological features of a simple feedforward neural network's layer representations of a modified torus with a Klein bottle-like twist were computed. The network appeared to approximate homeomorphisms in early layers, before significantly changing the topology of the data in deeper layers. The resulting noise hampered the ability of persistent homology to compute these features, however similar topological features seemed to persist longer in a network with a bijective activation function.
\end{abstract}

\section{Introduction}

In recent years, deep neural networks have revolutionized many computing problems once thought to be difficult. However, they are often referred to as black boxes and the mechanisms through which they learn have remained elusive. Insight into how a neural network internally represents the dataset it is trained on will provide understanding into what makes for effective training data, and how a neural network extracts relevant features from a dataset.

\subsection{Layer Representations}

A feedforward neural network, $N$, that acts on input data $x$ can be viewed as a composition of $n$ functions, where each function represents a layer, $L_i$,

$$N(x)=L_n(L_{n-1}(\dots L_1(x)))$$

Each layer, acting on input $z$, consists of a weights matrix $W$, bias vector $b$, and (usually) nonlinear activation function $\sigma$,

$$L_i(z)=\sigma(Wz+b)$$

The output of $L_i$ is what we will refer to as a layer representation. It is important to note that more complicated architectures are not restricted to a linear chain structure and thus can have skip and recurrent connections that complicate their graph structure. However, skip connections simply take in multiple inputs and recurrent (and recursive) connections can be unrolled and layer representations can be recovered.

\subsection{Overview of Paper}

Section 2 gives relevant background about topology and its relation to neural networks. Section 3 gives an overview of persistent homology, a technique to compute the topological features of a point-cloud dataset. Section 4 outlines the experiment performed and its results. Finally, Section 5 has the discussion of results and directions for future work.

\section{Background and Previous Work}

We will define topology, give motivation for the homology groups of a topological space, its applicability to neural networks, and review previous work on the subject.

\subsection{Definition of Topology}

Topology is the study of geometric objects with a particular structure that allows for a rigorous treatment of the concepts of "bending" and "twisting" a space and how properties of the space are preserved under these transformations. Formally, a topological space is the tuple $(X,\tau)$, where $X$ is some set and $\tau$ is a multiset consisting of subsets of $X$. Armstrong (1983) gives the following definition of a topological space,

\begin{definition}[Topological space]
\label{topology}
Given a set $X$ and a multiset $\tau$ that consists of subsets of $X$, a topological space is the tuple $(X,\tau)$ that fulfills the following axioms,

\begin{enumerate}
    \item $\emptyset\in\tau$ and $X\in\tau$
    \item $\forall S_i\in\tau, \bigcup_iS_i\in\tau$ for finite or infinite unions
    \item $\forall S_i\in\tau, \bigcap_iS_i\in\tau$ for only finite intersections
\end{enumerate}

\end{definition}

Any member of $\tau$ is termed an open set (and its complement is a closed set), and $\tau$ is the topology on $X$. 

Another central concept in topology is continuous deformation between topological spaces, and this is formalized in the idea of a homeomorphism, defined as,

\begin{definition}[Homeomorphism]
\label{homeomorphism}
A homeomorphism $f:X\to Y$ is an isomorphism between two topological spaces $X$ and $Y$ and therefore fulfills the following critera,

\begin{enumerate}
    \item $f$ is bijective
    \item $f$ is continuous
    \item $f^{-1}$ is continuous
\end{enumerate}

\end{definition}

\subsection{Homology}

The homology of a topological space informally characterizes the number of "holes" in the space. If we take a cycle to be a generalization of a closed loop on some space, such as on the surface of the sphere $S^2$, we can classify cycles by whether or not they can be continuously deformed into each other. If two cycles cannot be deformed into each other, it is said that there exists a hole on the topological space. 

A 0-dimensional hole is a connected component, a 1-dimensional hole is a loop, a 2-dimensional hole is a shell, and so on. The study of these holes requires a formal description of the boundaries on topological spaces. Boundaries in general are linear combinations of more basic geometric objects, which motivates us to introduce some sort of structure to allow for this. The homology of a topological space $X$ is formally represented by its homology groups,

$$H_0(X),H_1(X),H_2(X),\dots$$

Each homology group $H_k(X)$ has an abelian group structure, and as such we naturally use $\mathbb{Z}$. We refer to the rank of a homology group as the Betti number $b_k$. In general,

$$H_k(X)=\mathbb{Z}\times\cdots\times\mathbb{Z}=Z_{b_k}$$

A Betti number counts the number of holes in a topological space and is topologically invariant, which makes it an ideal candidate to judge the topological features of a space. Under homeomorphism, the Betti numbers of our domain and codomain remain unchanged. It is important to note that the Betti numbers do not account for all topological invariants, such as torsion. Torsion refers to features such as the twist of a Möbius strip. 

The Betti numbers of $S^1$, the circle, are $b_0=1,b_1=1$, and 0 otherwise. This is because there is one connected component, and one loop (1-dimensional boundary). The Betti numbers of $S^2$ are $b_0=1,b_2=1$, and 0 otherwise. This is because there is 1 connected component and one 2-dimensional boundary (around the interior of the sphere). Notably, there are no loops that can be drawn on the surface that cannot be deformed to a single point. The torus $T^2$, a doughnut-shaped object, has Betti numbers $b_0=1,b_1=2,b_2=1$, and 0 otherwise. There is one connected component, and two loops (one around the ring and another around the "main hole"). Lastly, there is a 2-dimensional boundary around the interior. It is interesting to note $T^2=S^1\times S^1$.

\subsection{Application to Neural Networks}

We are interested in whether the layer representation $L_i$ acts similar to a homeomorphism in that it allows a neural network to represent the topological features of a dataset. Unfortunately, neural network layers are not in general homeomorphisms,.

\begin{theorem}
A neural network layer, $L(x)=\sigma(Wx+b)$, need not be a homeomorphism.
\begin{proof}
If $W$ is not a member of the general linear group $\mathrm{GL}_n(\mathbb{R})$, it is not invertible and therefore no homeomorphism exists by \ref{homeomorphism}. Additionally, if $\sigma$ is not bijective, no homeomorphism can exist also by \ref{homeomorphism}.
\end{proof}
\end{theorem}

However, the topological approach is still useful. If layer representations are examined and are found to resemble the topological features of the dataset, this suggests a reason neural networks are effective is that they learn a robust representation of the topological features of a space. 

\subsection{Previous Work}

Studying neural network learning with topology has been a popular approach in theoretical machine learning. This is best exemplified with the concept of the manifold hypothesis. Fefferman, et al. (2013) gives the following definition of the manifold hypothesis,

\begin{definition}[Manifold Hypothesis]
\label{manifold}
High-dimensional data tends to lie near low-dimensional manifolds.
\end{definition}

This concept is best explained with an example. Imagine there exists a model to classify $m\times n$ images between cats and dogs. The data space is $\mathbb{R}^{mn}$. However, this space clearly contains data not relevant to the task at hand (such as images of flowers or random noise), and the manifold hypothesis conjectures that there exists a much lower-dimensional submanifold of the data space that approximates the relevant data.

Given that manifolds are topological spaces, it is reasonable to assume methods from topology can be used to gain insight into the learning process, should the manifold hypothesis be correct. There exists a field called topological data analysis (TDA) precisely focused on extracting topological features from data. The most common technique used in TDA is persistent homology. Persistent homology will be described in the next section, but it essentially computes whether the features corresponding to each homology group exist in a point-cloud, e.g. if one can draw loops via linear combinations of  datapoints. Montúfar, et al. (2020) showed that a neural network can be trained to approximate topological features similar to persistent homology. While this paper will use persistent homology to compute topological features, it is encouraging that a neural network can be trained to compute the topological features of a dataset.

\section{Persistent Homology}

Persistent homology is a technique to compute the topological features of a point-cloud dataset and returns these results in a persistence diagram.

\subsection{Method}

Given a point-cloud, persistent homology grows $n$-dimensional balls around each point and marks when these balls intersect with the balls of another point. By connecting these balls together, one can represent the data as simplicial complexes by chaining together simplices (generalizations of points, lines, and triangles in higher-dimensional space). The validity of this is a result of the abelian group structure imposed on the homology of our dataset. These simplicial complexes can then represent topological features such as loops. At some point, all points will intersect as the balls will grow to encompass the whole dataset. We record the birth and death times of these features and plot them in a persistence diagram.

\subsection{Persistence Diagrams}

We generated a variation of a torus with a Klein bottle-like twist, $X$, according to the following parametrization,

$$x = (R + 2P\cos\theta)\cos\phi\;,$$
$$y = (R + 2P\cos\theta)\sin\phi\;,$$
$$z = 2P\sin\theta\cos\frac{\phi}{2}\;,$$
$$w=2P\sin\theta\sin\frac{\phi}{2}$$

We display the point-cloud (only the first three dimensions) and its persistence diagram in figure 1. For $H_0(X)$, we can see that all features but one die soon after they are birthed, which is indicated by points residing close to the diagonal of the diagram. The single point that persists at infinity corresponds to the single connected component obtained when all the balls intersect in persistent homology. For $H_1(X)$, we similarly see most features die close to birth. However, there are two features that appear to persist for a significant time before ultimately dying during their merger. These correspond to loops found by chaining together simplices generated from datapoints to form a simplicial complex by the algorithm. In general, the Betti numbers are difficult to extract for complicated persistence diagrams, however we can make a reasonable guess that were this manifold to be taken in the continuum limit, we would get Betti numbers $b_0=1$ and $b_1=2$.

\begin{figure}
\begin{subfigure}{0.45\textwidth}
  \centering
  \includegraphics[width=\textwidth]{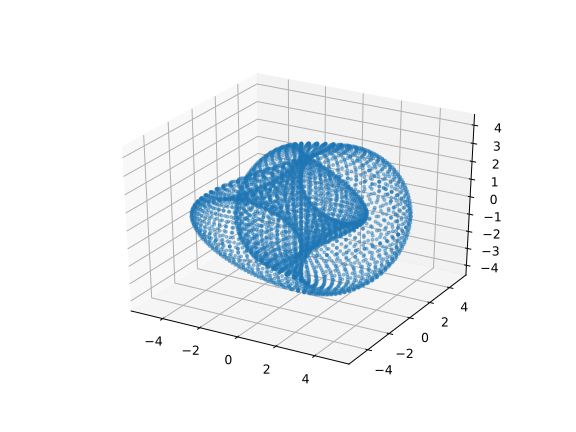}
  \caption{The three-dimensional projection of the point cloud.}
\end{subfigure}
\hfill
\begin{subfigure}{0.45\textwidth}
  \centering
  \includegraphics[width=\textwidth]{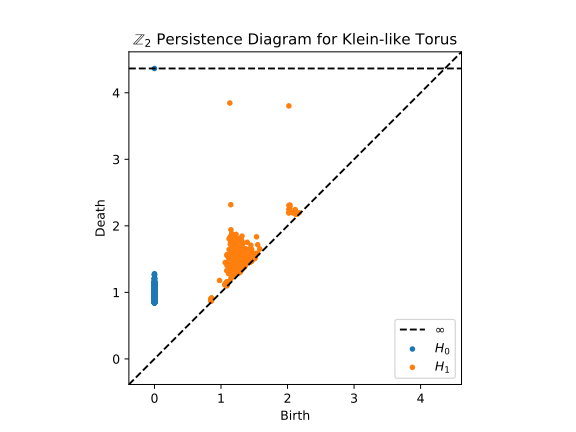}
  \caption{The persistence diagram of the point-cloud.}
\end{subfigure}
\caption{A torus with a Klein bottle-like twist along with its persistence diagram.}
\end{figure}

\section{Experiment and Results}

\begin{table}
  \caption{Architecture of Neural Network}
  \centering
  \begin{tabular}{lll}
    \toprule
    Name     & Hidden Units     & Activation Function \\
    \midrule
    Input & 4  & ReLU    \\ 
    Layer 1     & 10 & ReLU      \\ 
    Layer 2     & 30      & ReLU  \\ 
    Layer 3     & 10            & ReLU \\ 
    Output & 1 & Sigmoid \\
    \bottomrule
  \end{tabular}
\end{table}

This section outlines the experiment conducted, the layer representations obtained, their persistence diagrams, and a PCA projection of them. 

\subsection{Experiment Design}

The experiment consisted of the modified torus parametrized in the previous section along with noise sampled from the uniform distribution. The total dataset length was 9800. The network, whose architecture is given in table 1, was trained on the binary classification task of distinguishing whether or not a point lay on the modified torus. The model was built using TensorFlow.

Using the Adam optimizer, the network was trained for 300 epochs when an accuracy of approximately 98\% was obtained. Using the HDBSCAN clustering algorithm, the output of each layer was clustered and projected back onto the data space. Persistence diagrams on the clusters were calculated using Ripser. Lastly, principal component analysis (PCA) was applied on the layer representations corresponding to the torus to project them onto three dimensions.

The experiment was then repeated using the Tanh activation function on the network architecture described in table 1.

\subsection{Results}

The results from the experiments are summarized in figures 2 to 13. We notice that the clustering becomes progressively less defined throughout the layers as the network's representation becomes more consolidated. Notably, despite the visual resemblance to the data, persistent homology appeared unable to calculate topological features with the same degree of accuracy as on the raw data. The PCA projections obtained appear to resemble the original data less as the layers become deeper. It is noted that we were unable to compute the persistence diagrams of the clusterings of the third layer of the ReLU network due to the these clusterings being extremely noisy.

\begin{figure}
\begin{subfigure}{0.45\textwidth}
  \centering
  \includegraphics[width=\textwidth]{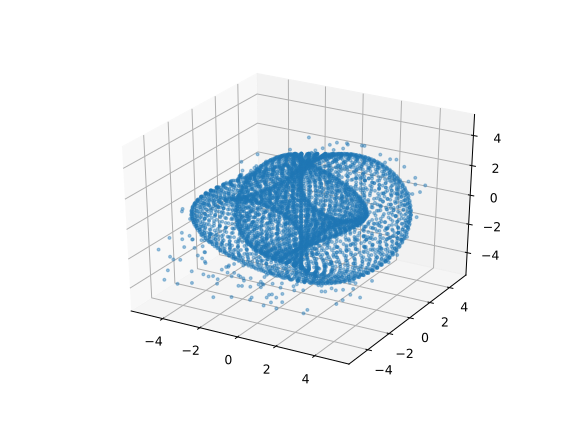}
\end{subfigure}
\hfill
\begin{subfigure}{0.45\textwidth}
  \centering
  \includegraphics[width=\textwidth]{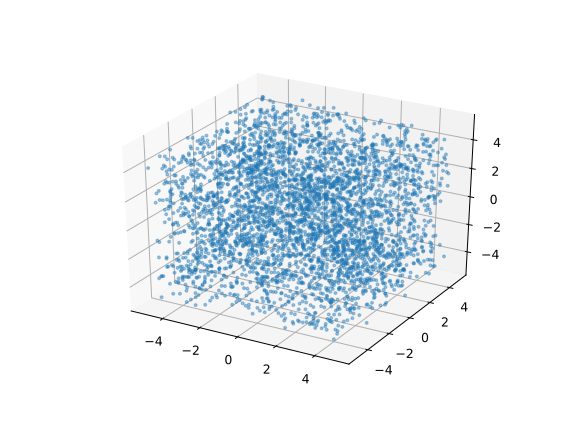}
\end{subfigure}
\caption{First two clusters of the layer 1 representations of the ReLU network.}
\end{figure}

\begin{figure}
\begin{subfigure}{0.45\textwidth}
  \centering
  \includegraphics[width=\textwidth]{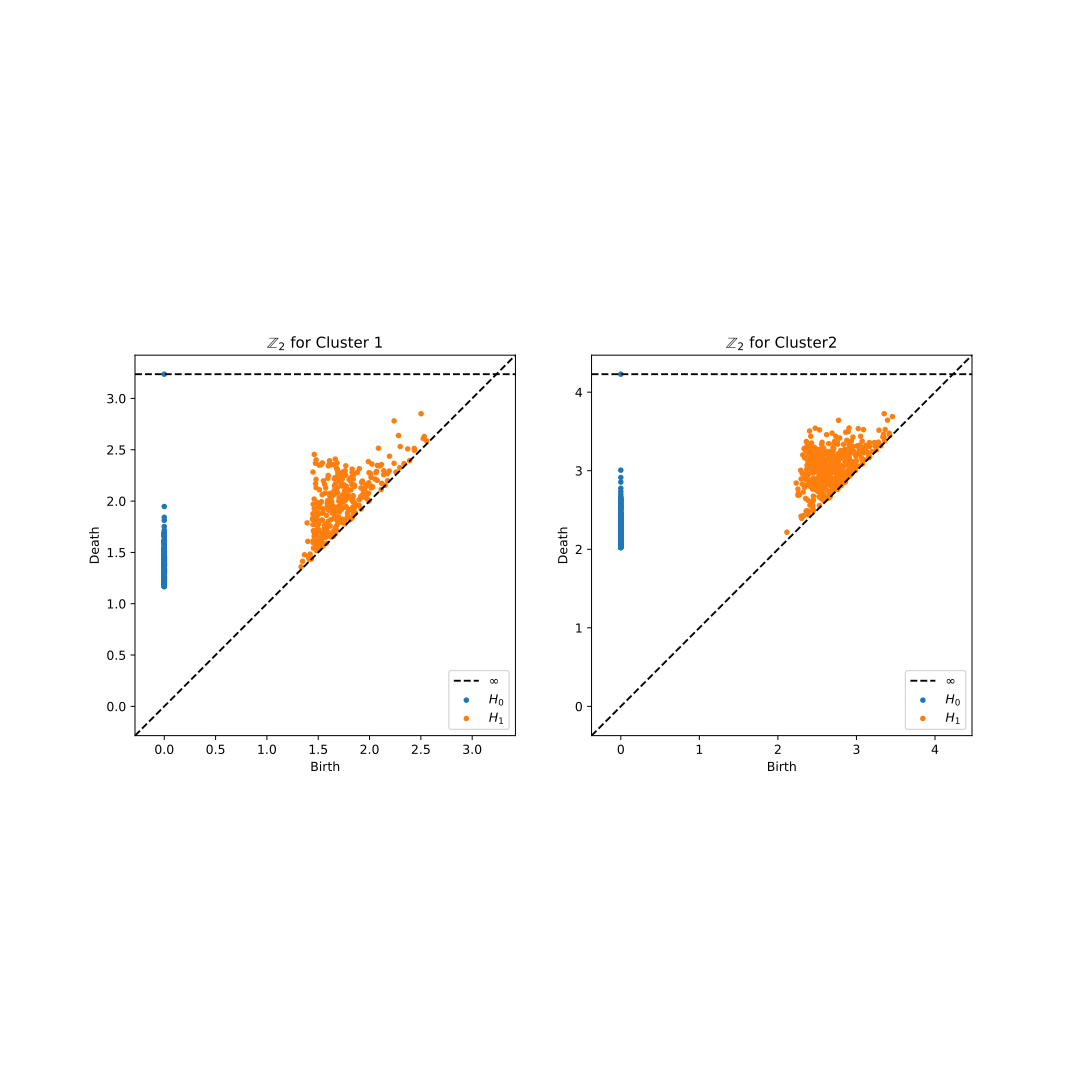}
  \caption{Persistence diagrams of clusters.}
\end{subfigure}
\hfill
\begin{subfigure}{0.45\textwidth}
  \centering
  \includegraphics[width=\textwidth]{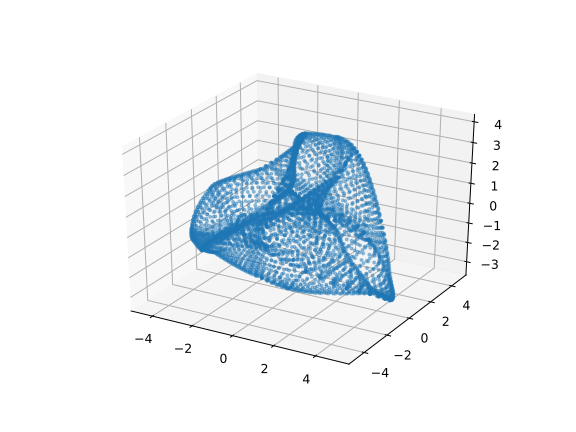}
  \caption{PCA projection of torus.}
\end{subfigure}
\caption{Persistence diagrams of layer 1 clusters for the ReLU network and PCA projection.}
\end{figure}

\begin{figure}
\begin{subfigure}{0.45\textwidth}
  \centering
  \includegraphics[width=\textwidth]{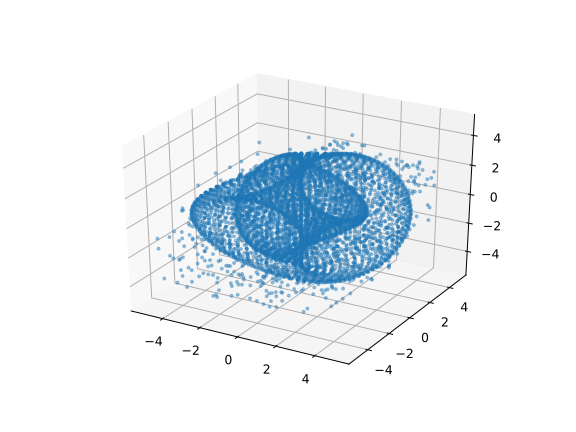}
\end{subfigure}
\hfill
\begin{subfigure}{0.45\textwidth}
  \centering
  \includegraphics[width=\textwidth]{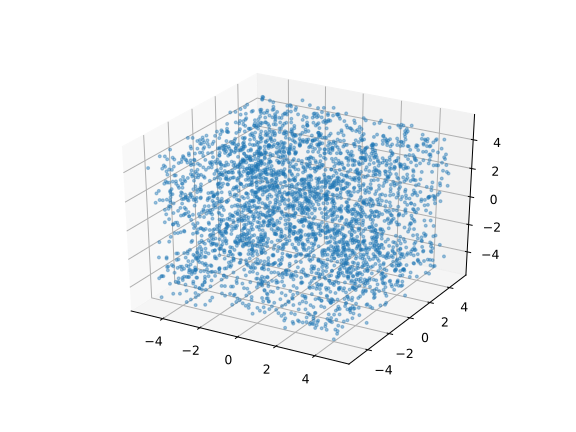}
\end{subfigure}
\caption{First two clusters of the layer 2 representations of the ReLU network.}
\end{figure}

\begin{figure}
\begin{subfigure}{0.45\textwidth}
  \centering
  \includegraphics[width=\textwidth]{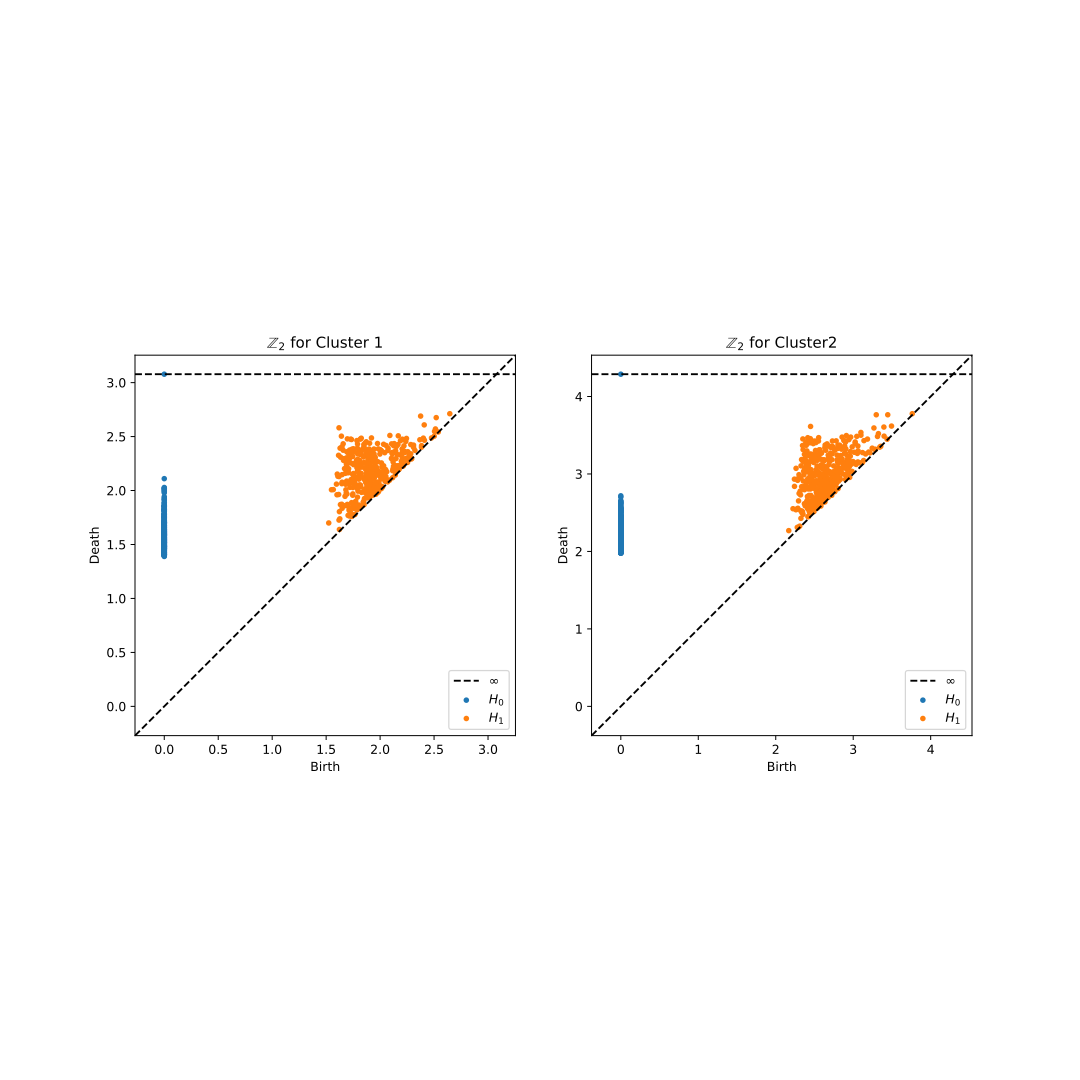}
  \caption{Persistence diagrams of clusters.}
\end{subfigure}
\hfill
\begin{subfigure}{0.45\textwidth}
  \centering
  \includegraphics[width=\textwidth]{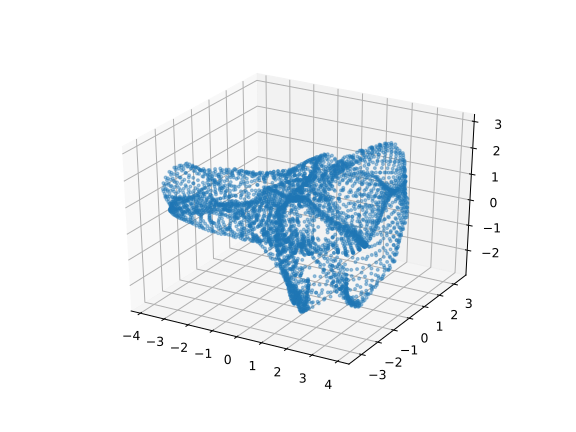}
  \caption{PCA projection of torus.}
\end{subfigure}
\caption{Persistence diagrams of layer 2 clusters for the ReLU network and PCA projection.}
\end{figure}

\begin{figure}
\begin{subfigure}{0.45\textwidth}
  \centering
  \includegraphics[width=\textwidth]{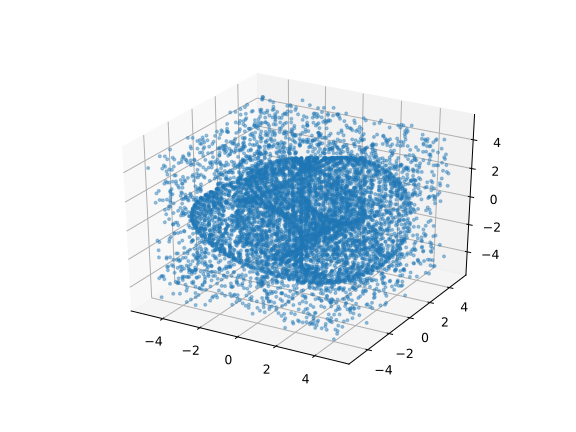}
\end{subfigure}
\hfill
\begin{subfigure}{0.45\textwidth}
  \centering
  \includegraphics[width=\textwidth]{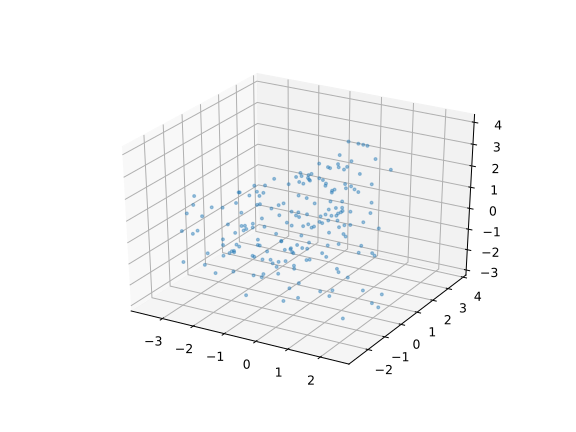}
\end{subfigure}
\caption{First two clusters of the layer 3 representations of the ReLU network.}
\end{figure}

\begin{figure}
  \centering
  \includegraphics[width=\textwidth]{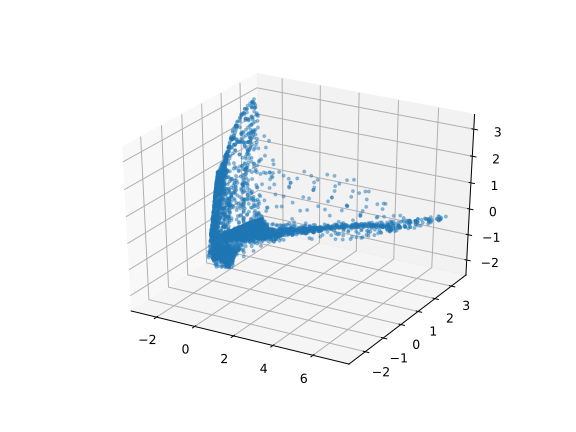}
  \caption{PCA projection of the layer 3 representation of the ReLU network.}
\end{figure}

\begin{figure}
\begin{subfigure}{0.45\textwidth}
  \centering
  \includegraphics[width=\textwidth]{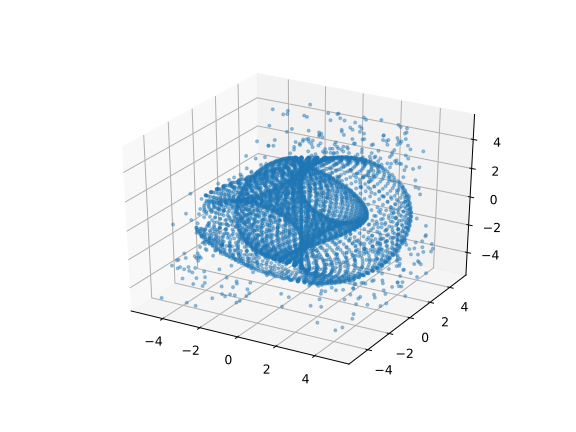}
\end{subfigure}
\hfill
\begin{subfigure}{0.45\textwidth}
  \centering
  \includegraphics[width=\textwidth]{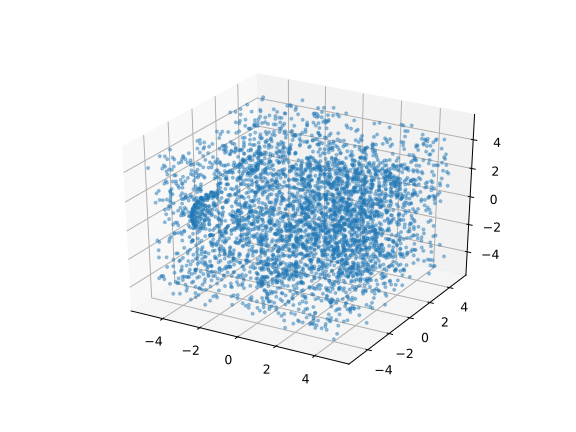}
\end{subfigure}
\caption{First two clusters of the layer 1 representations of the Tanh network.}
\end{figure}

\begin{figure}
\begin{subfigure}{0.45\textwidth}
  \centering
  \includegraphics[width=\textwidth]{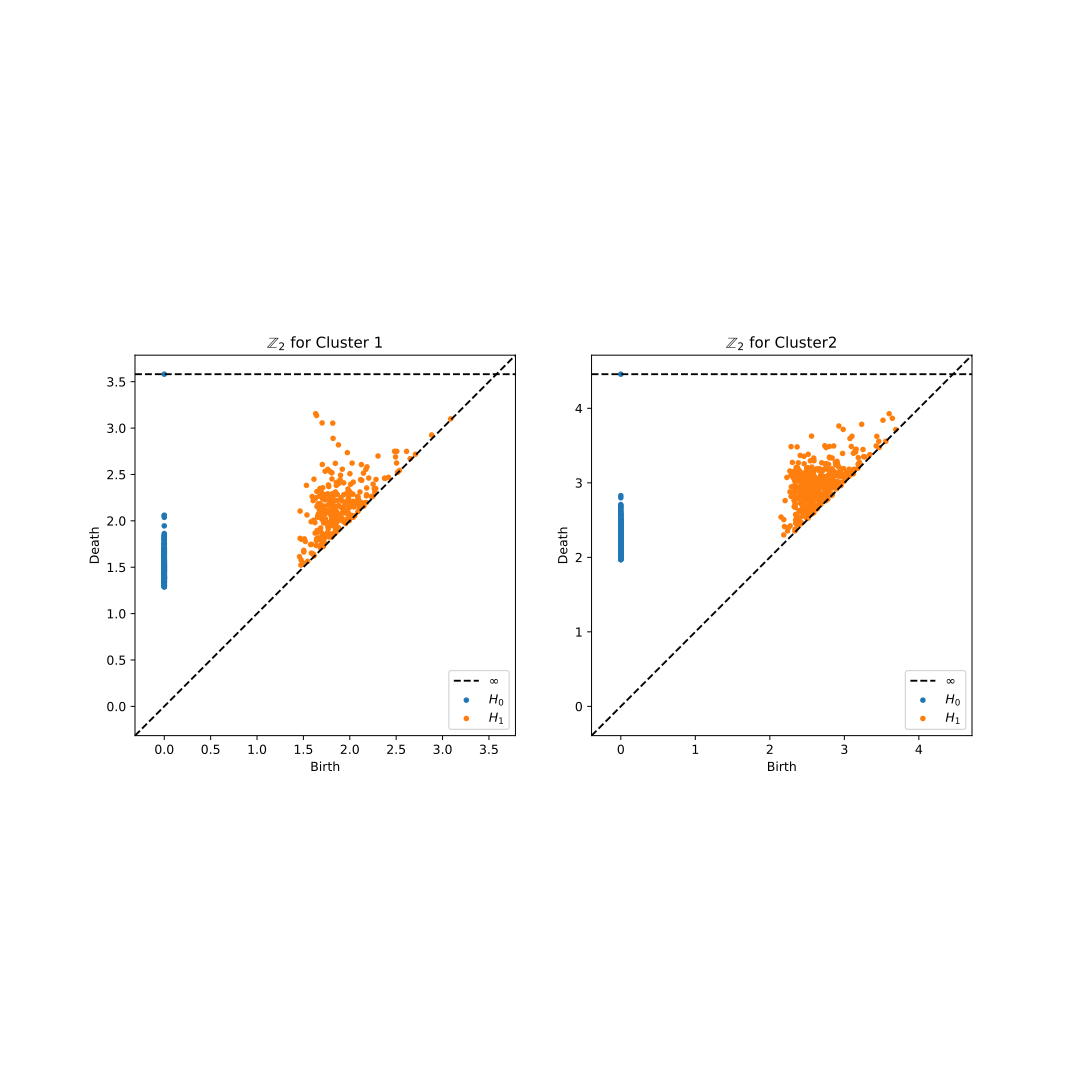}
  \caption{Persistence diagrams of clusters.}
\end{subfigure}
\hfill
\begin{subfigure}{0.45\textwidth}
  \centering
  \includegraphics[width=\textwidth]{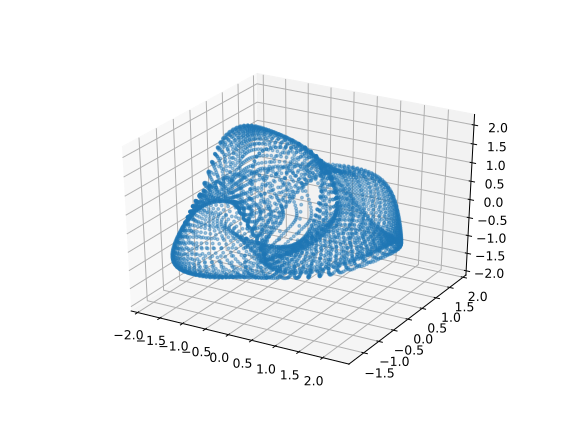}
  \caption{PCA projection of torus.}
\end{subfigure}
\caption{Persistence diagrams of layer 1 clusters for the Tanh network and PCA projection.}
\end{figure}

\begin{figure}
\begin{subfigure}{0.45\textwidth}
  \centering
  \includegraphics[width=\textwidth]{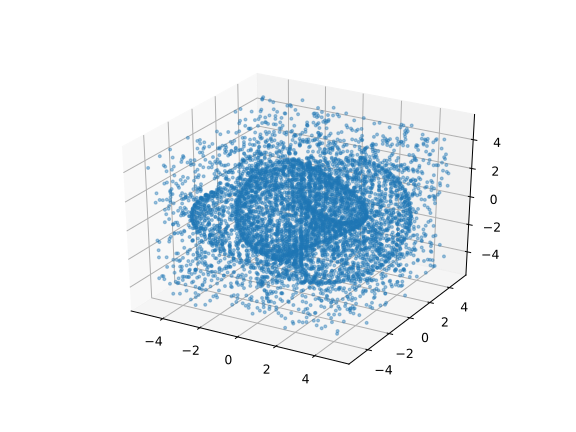}
\end{subfigure}
\hfill
\begin{subfigure}{0.45\textwidth}
  \centering
  \includegraphics[width=\textwidth]{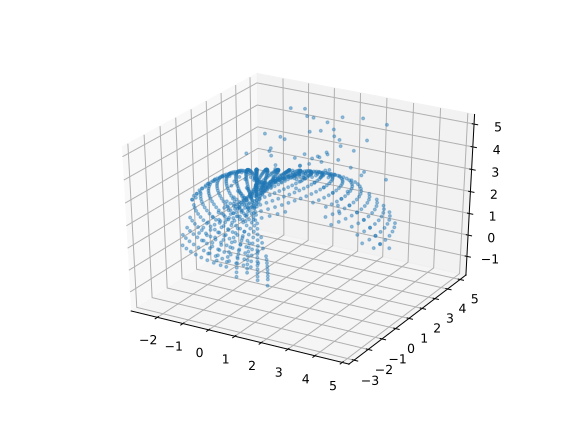}
\end{subfigure}
\caption{First two clusters of the layer 2 representations of the Tanh network.}
\end{figure}

\begin{figure}
\begin{subfigure}{0.45\textwidth}
  \centering
  \includegraphics[width=\textwidth]{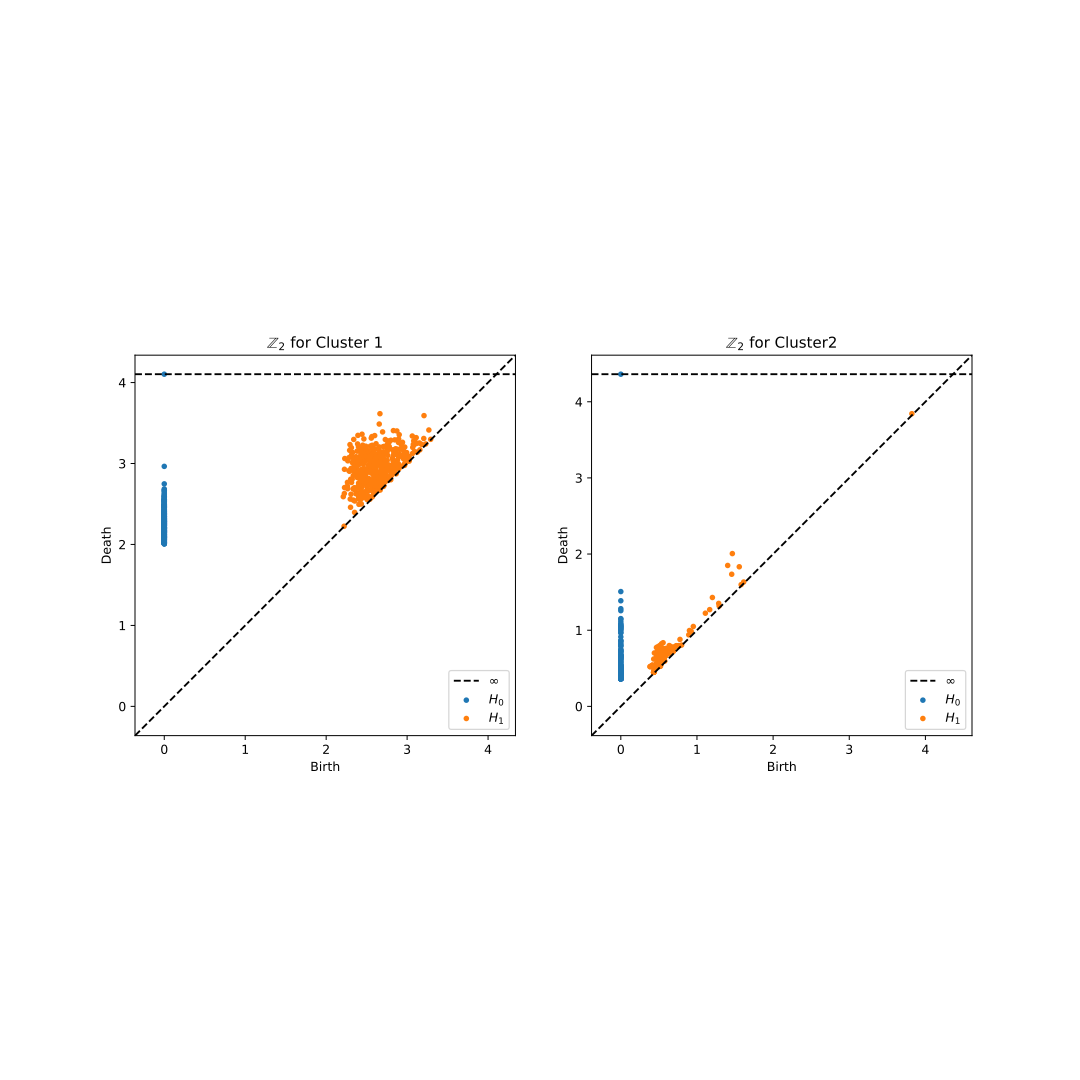}
  \caption{Persistence diagrams of clusters.}
\end{subfigure}
\hfill
\begin{subfigure}{0.45\textwidth}
  \centering
  \includegraphics[width=\textwidth]{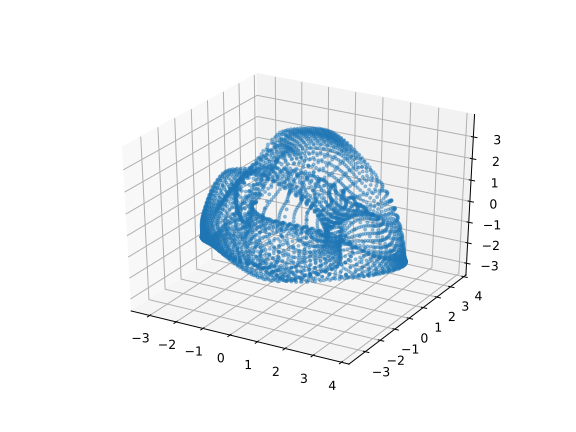}
  \caption{PCA projection of torus.}
\end{subfigure}
\caption{Persistence diagrams of layer 2 clusters for the Tanh network and PCA projection.}
\end{figure}

\begin{figure}
\begin{subfigure}{0.45\textwidth}
  \centering
  \includegraphics[width=\textwidth]{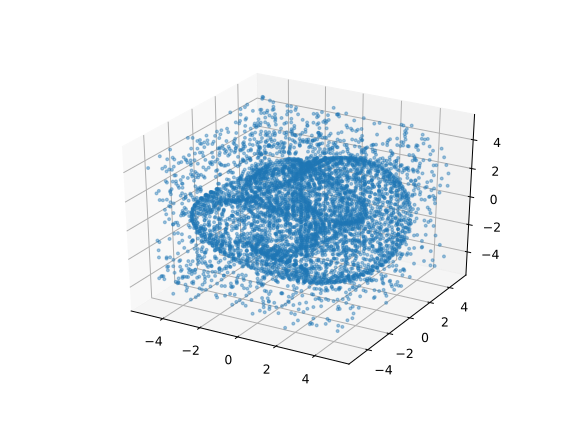}
\end{subfigure}
\hfill
\begin{subfigure}{0.45\textwidth}
  \centering
  \includegraphics[width=\textwidth]{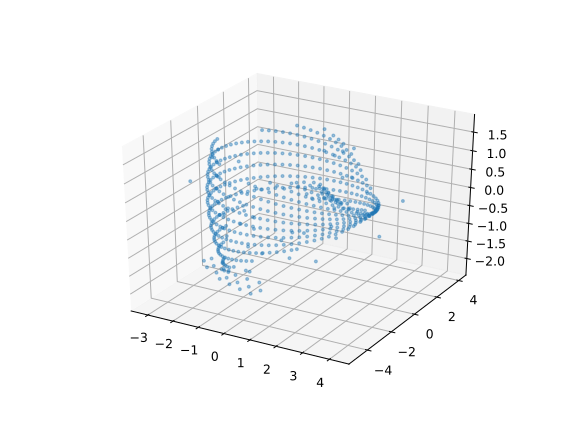}
\end{subfigure}
\caption{First two clusters of the layer 3 representations of the Tanh network.}
\end{figure}

\begin{figure}
\begin{subfigure}{0.45\textwidth}
  \centering
  \includegraphics[width=\textwidth]{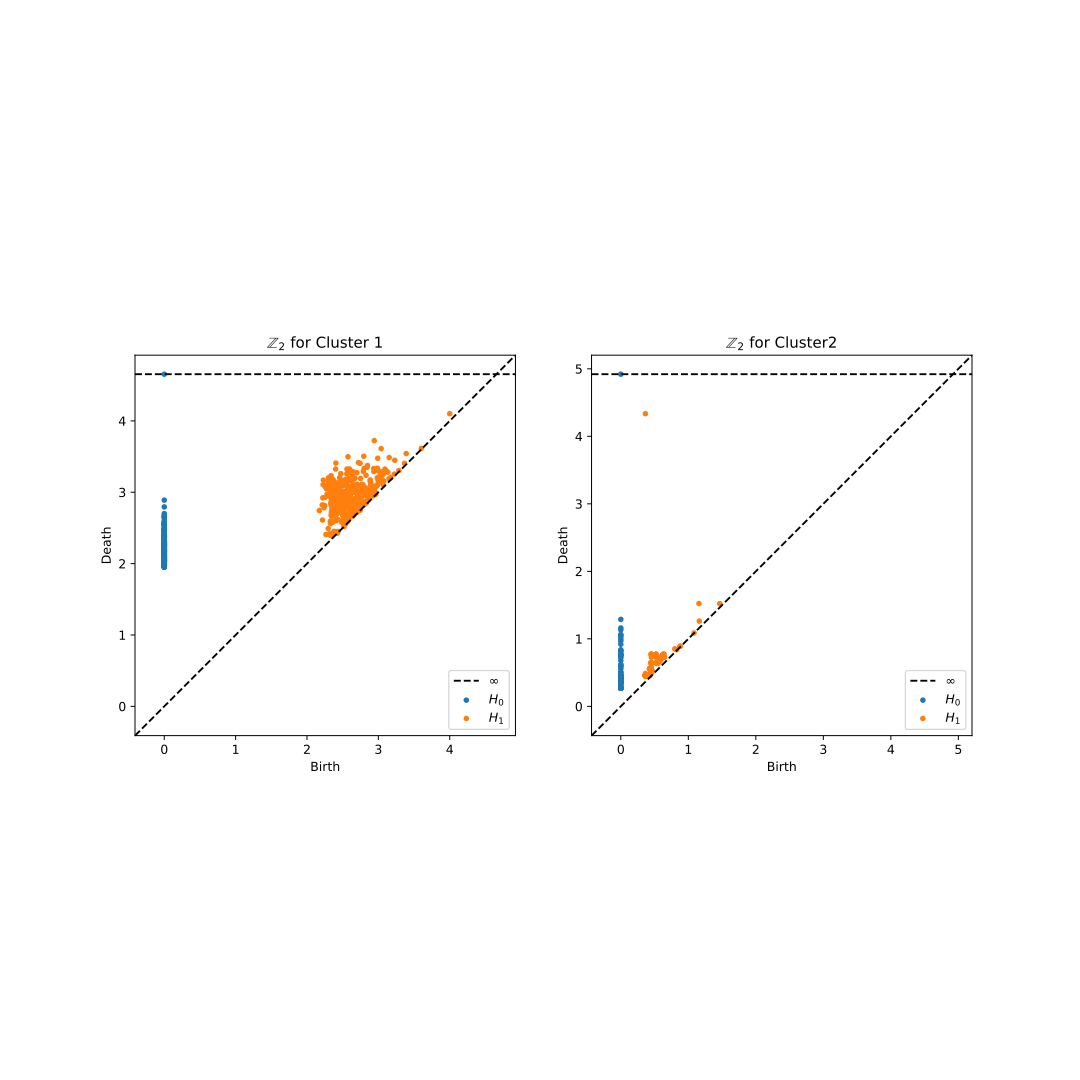}
  \caption{Persistence diagrams of clusters.}
\end{subfigure}
\hfill
\begin{subfigure}{0.45\textwidth}
  \centering
  \includegraphics[width=\textwidth]{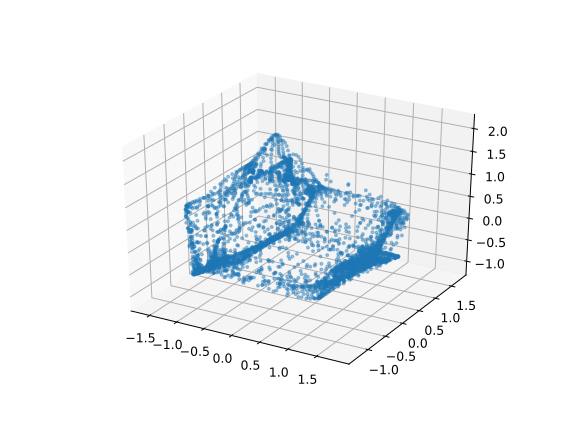}
  \caption{PCA projection of torus.}
\end{subfigure}
\caption{Persistence diagrams of layer 3 clusters for the Tanh network and PCA projection.}
\end{figure}

\section{Discussion and Further Work}

The visual results from the experiments seem to indicate that the network is approximating homeomorphisms in early layers, before deeper layer representations obtain a topology markedly different than the original data. This could result from repeated application of non-homeomorphic layers making recovery of the underlying topology difficult. This appears to be correlated with inability of HDBSCAN to properly cluster the layer representations of deeper layers. However, we notice that while not as defined as the raw data, the persistence diagrams indicate the presence of topological features similar to the original data for the first two layers of both networks, albeit with a much shorter lifetime due to interference from the noise of the clusterings. This brings up the possibility that the network is attempting to isolate the relevant topological features for classification, and excising superfluous features. This would appear to explain the progressively sparser PCA projections obtained as the layers get deeper. The deep PCA projections seem to cluster points in more localized areas, leaving more empty portions in the space unlike the spread-out representations of earlier layers.

Furthermore, the PCA projections seem to indicate that the network approximates a continuous deformation for the first two layers of the Tanh network and first layer of the ReLU network, before the topology becomes unrecognizable. This could a result of ReLU being a surjective function, rather than bijective function like Tanh, making it harder for the layer to approximate a homeomorphism. 

It is noted that nothing in this study is conclusive, and these are observations gathered from a preliminary experiment. Further work must be done in testing neural networks on more varied topologies and perhaps using alternative clustering algorithms as the noise included in these clusters appear to disrupt persistent homology.

\section*{References}

{
\small

[1] Abadi, M. et al. (2015). TensorFlow: Large-Scale Machine Learning on Heterogenous Systems. https://doi.org/10.5281/zenodo.4724125.

[2] Armstrong, M. A. (1983). Basic Topology. \textit{Springer}, doi:10.1007/978-1-4757-1793-8.

[3] Fefferman, C., Mitter, S., \& Narayanan, H. (2013). Testing the Manifold Hypothesis. 
\textit{arXiv}, https://doi.org/10.48550/arXiv.1310.0425.

[4] McInnes, L., Healy, J., \& Astels, S. (2017). hdbscan: Hierarchical density based clustering. \textit{The Journal of Open Source Software}, 2. https://doi.org/10.21105/joss.00205.

[5] Montúfar, G., Otter, N., \& Wang, Y. (2020). Can neural networks learn persistent homology features?. \textit{arxiv}, https://doi.org/10.48550/arXiv.2011.14688.

[6] Tralie et al. (2018). Ripser.py: A Lean Persistent Homology Libary for Python. \textit{Journal of Open Software}, 3(29), 925, https://doi.org/10.21105/joss.00925.

}

\end{document}